\documentclass{article}
\usepackage{kr}
\usepackage[inline]{enumitem}
\usepackage{times}
\usepackage{soul}
\usepackage{url}
\usepackage[hidelinks]{hyperref}
\usepackage[utf8]{inputenc}
\usepackage[small]{caption}
\usepackage{graphicx}
\usepackage[fleqn]{amsmath}
\usepackage{amsthm}
\usepackage{amssymb}
\usepackage{booktabs}
\usepackage{algorithm}
\usepackage{algorithmic}
\usepackage{verbatim}
\usepackage{mathabx}
\usepackage[hang,flushmargin]{footmisc}
\usepackage{subcaption}
\newtheorem{example}{Example}
\newtheorem{theorem}{Theorem}

\title{Commonsense Spatial Reasoning for Visually Intelligent Agents}

\author{%
Agnese Chiatti$^1$\and
Gianluca Bardaro$^1$\and
Enrico Motta$^1$\and
Enrico Daga$^1$ \\
\affiliations
$^1$Knowledge Media Institute, The Open University, United Kingdom\\

\emails
\{name.surname\}@open.ac.uk
}

\begin{document}

\maketitle

\begin{abstract}
Service robots are expected to reliably make sense of complex, fast-changing environments. From a cognitive standpoint, they need the appropriate reasoning capabilities and background knowledge required to exhibit human-like Visual Intelligence. In particular, our prior work has shown that the ability to reason about spatial relations between objects in the world is a key requirement for the development of Visually Intelligent Agents. In this paper, we present a framework for commonsense spatial reasoning which is tailored to real-world robotic applications. Differently from prior approaches to qualitative spatial reasoning, the proposed framework is robust to variations in the robot's viewpoint and object orientation. The spatial relations in the proposed framework are also mapped to the types of commonsense predicates used to describe typical object configurations in English. In addition, we also show how this formally-defined framework can be implemented in a concrete spatial database. 
\end{abstract}

\section{Introduction}
 In all cases where it is inconvenient or even dangerous for us to intervene, there is an incentive to delegate tasks to \textit{service robots} - or \textit{robot assistants}: e.g., under the extreme conditions imposed by space explorations \cite{nilsson2018toward}, in hazardous manufacturing environments \cite{liu2020remote}, or whenever social distance needs to be maintained \cite{yang2020keep}. Before delegating complex tasks to robots, however, we need to ensure that they can reliably \textit{make sense} of the stimuli coming from their sensors. Autonomous sensemaking remains an open challenge, because it requires not only to reconcile the high-volume and diverse data collected from real-world settings \cite{alatise2020review}, but also to actually understand these data, going beyond mere pattern recognition \cite{lake_building_2017,davis_commonsense_2015}. 

From a vision perspective, the problem of robot sensemaking becomes one of enhancing the \textit{Visual Intelligence} of service robots, i.e., their ability to make sense of the environment through their vision system and epistemic competences \cite{chiatti_towards_2020}. 
Naturally, several epistemic competences are required to build \textit{Visually Intelligent Agents (VIA)}. For instance, let us consider the case of HanS, a Health \& Safety robot inspector. HanS is expected to autonomously detect potentially threatening situations, such as the fire hazard posed by a sweater left to dry on top of an electric heater. To assess the risk associated with this situation, HanS first needs to recognise the sweater and the heater in question, i.e., it needs to exhibit robust \textit{object recognition} capabilities. It also needs \textit{spatial reasoning} capabilities, to infer that the sweater is touching the heater. Moreover, it also needs to know that sweaters are made of cloth and that a piece of cloth clogging an electric radiator can catch fire. The list goes on. 

In \cite{chiatti_towards_2020}, we identified a framework of \textit{epistemic requirements}, i.e., knowledge properties and reasoning capabilities which are needed to develop \textit{Visually Intelligent Agents (VIA)}. To form hypotheses on which epistemic requirements are more likely to significantly enhance the Visual Intelligence of a robot, we also mapped these epistemic ingredients to the types of object classification errors emerging from one of HanS' scouting routines. This error analysis highlighted that the majority of misclassifications could in principle have been avoided, if the robot was capable of considering: (i) the canonical size of objects, e.g., that mugs are generally smaller than bins, as well as (ii) the typical \textit{Qualitative Spatial Relations (QSR)} between objects. For instance, a fire extinguisher may be mistaken for a bottle due to its shape. However, the proximity of a fire extinguisher sign is a strong indication that the observed object is in fact a fire extinguisher. This element of \textit{typicality} relates to the broader objective of developing AI systems which can reason about what is \textit{plausible} \cite{davis_commonsense_2015}, i.e., which exhibit \textit{common sense} \cite{levesque_common_nodate} and \textit{Intuitive Physics} reasoning abilities \cite{hayes_second_1988,lake_building_2017}. Our most recent findings \cite{chiatti2021aaaimake} confirmed that combining state-of-the-art Machine Learning methods with a component able to reason about object sizes improves the robot's object recognition performance. In this paper, we progress this line of research by characterising commonsense QSR between objects.          

The problem of representing spatial relations has been actively researched for decades, producing many theoretical frameworks for autonomous spatial reasoning \cite{cohn_chapter_2008}. In robotics, \textit{semantic mapping} \cite{nuchter2008towards,kostavelis2015semantic} and \textit{object anchoring} methods \cite{coradeschi2003introduction} have enabled linking the robot sensor data and symbolic knowledge to the geometric maps modelling its environment. To combine the best of both worlds, a number of approaches \cite{deeken_grounding_2018,kunze_combining_2014,young_semantic_2017} have linked the spatial representations within semantic maps to the higher-level formal definitions provided by AI theories. 
In this paper, we propose a novel spatial reasoning framework which extends the work in \cite{deeken_grounding_2018,borrmann_query_2010}, to account for variations in the robot's viewpoint and in the relative orientation of objects. Moreover, we formally map the Qualitative Spatial Relations composing this framework to the type of linguistic predicates used to describe \textit{commonsense spatial relations} in English, which are discussed within seminal theories of spatial cognition \cite{landau_what_1993,herskovits_language_1997}. Finally, we show how the proposed framework can be implemented in state-of-the-art Geographic Information Systems (GIS), to support commonsense spatial reasoning in real-world robotic scenarios.

\section{Related work}
Broadly speaking, spatial relations can be represented qualitatively - e.g., A contains B - or quantitatively - e.g., the angle between A and B is $\theta$ \cite{thippur_comparison_2015}. Following \cite{borrmann_query_2010}, Qualitative Spatial Relations (QSR) can be further characterised as (i) \textit{metric}, i.e., based on the metric distance between objects (ii) \textit{topological}, i.e., describing the neighbourhood of objects, and (iii) \textit{directional}, i.e., relative to the axis directions in a reference coordinate system. The interested reader is referred to \cite{cohn_chapter_2008} for a foundational review of qualitative spatial representations. Compared to quantitative representations, qualitative representations are more similar to the types of spatial predicates involved in natural language discourse. As a result, qualitative spatial representations are easier to interpret and aid Human-Robot Interaction \cite{sarthou2019semantic,sisbot_where_2019,thippur_non-parametric_2017}. Moreover, they are more similar to the types of linguistic predicates available within large-scale, general-purpose Knowledge Bases (KB), such as those surveyed in \cite{storks2019recent}. Crucially, they are also aligned with the spatial predicates provided with benchmark image collections for visual reasoning tasks, such as Visual Genome \cite{krishna2017visual} and SpatialSense \cite{yang_spatialsense_2019}. Thus, relying on qualitative representations has the potential to facilitate the repurposing of these resources in robotic contexts, especially given the paucity of comprehensive KBs for Visually Intelligent Agents \cite{chiatti_towards_2020}. 

Extensive efforts have been devoted to mapping the quantitative data collected through the robot's sensors to higher-level symbols describing a set of known object classes and their attributes \cite{nuchter2008towards,coradeschi2003introduction,kostavelis2015semantic}. These efforts have produced intermediate representational models also known as \textit{semantic maps}, i.e., maps that contain, ``in addition to spatial information about the environment, assignments of mapped features to entities of known classes'' \cite{nuchter2008towards}. 
Further approaches have been proposed, where the content of semantic maps is also interpreted with respect to formal theories of qualitative spatial reasoning \cite{young_semantic_2017,kunze_combining_2014,deeken_grounding_2018}. In general, spatial relations are expressed between object pairs, where one of the two objects is considered as a \textit{reference}, or \textit{landmark}: e.g., bike near house. \cite{young_semantic_2017} have used Ring Calculus to represent the closeness of objects.  \cite{kunze_combining_2014} have relied on ternary point calculus \cite{moratz2008qualitative} to model directional relations with respect to both the robot's location and the location of the reference object. Thus, the 3D regions occupied by objects are reduced to point-like objects on the 2D plane. Moreover, \cite{kunze_combining_2014} assumed that the robot's location does not change over time, and is always defined with respect to a tabletop. Differently from \cite{kunze_combining_2014}, \cite{deeken_grounding_2018} represented directional relations by comparing the 3D regions occupied by objects, through the halfspace-based model of \cite{borrmann_query_2010}. However, this model is based on the assumption that the robot's viewpoint is always aligned both with the global coordinate system of the map and with the inherent orientation of the observed objects. Thus, it is not suitable to model mobile robots making sense of the environment during navigation. In real-world scenarios, as the robot moves, its viewpoint changes over time and the objects observed will be oriented differently.  
Thus, we propose to combine the robot's viewpoint and the orientation of the reference object within a \textit{contextualised frame of reference}. This contextualised frame of reference allows us to define a contextualised 3D region, or \textit{Contextualised Bounding Box}, which represents the location of the object with respect to both the robot's viewpoint and the frame of reference of a landmark. Crucially, the contextualised frame of reference and Bounding Box can be defined for any combination of robot and landmark location, thus ensuring that this framework can scale to many real-world robotic scenarios.

\section{Proposed Framework}
To define a spatial reasoning framework which satisfies the concrete requirements of robot sensemaking, we extend the formal theory of spatial reasoning by \cite{borrmann_query_2010}. Moreover, we map the obtained spatial relations to the commonsense predicates used to describe spatial relations between objects in English. These predicates are gathered from cognitive theories \cite{landau_what_1993}. By making an explicit link between formal AI theories and informal linguistic representations, we obtain a framework for commonsense spatial reasoning in robotic scenarios. 

\subsubsection{Notation} In what follows, we model definitions as First Order Logic (FOL) statements. We represent logic variables through lowercase letters and constants through uppercase letters. We also use lowercase initials to denote functions, while uppercase initials symbolise predicates. For instance, \textit{sReg} is a function, whereas \textit{Above} is a predicate. Unless otherwise stated, free variables are universally quantified.  Finally, we use the standard notation (X, Y, Z) to denote reference axes, while $x,y,z$ are used to refer to the spatial coordinates with respect to those axes. 

\subsubsection{Spatial primitives} Our domain of discourse $\mathbb{D}$ is that of \textit{spatial objects}, i.e., physical objects, ``which have spatial extensions'' \cite{cohn_chapter_2008}. From this perspective, a spatial object is represented in terms of the associated \textit{spatial region}. In particular, our spatial primitive is the concept of \textit{spatial point}. Thus, spatial regions are represented as sets of spatial points, $p$. Let $P$ be the set of all spatial points, then, for each spatial object $o \in \mathbb{D}$, we assume the existence of a function $sReg$ which, given $o$, returns the subset of $P$ which includes all the points in the spatial region of $o$.  
\begin{align}
\textit{SpatialObj}(o) \Rightarrow \textit{sReg}(o) \subseteq P \\
\textit{SpatialObj}(o) \Rightarrow \textit{sReg}(o) \neq \emptyset
\end{align}
In particular, our focus is not on arbitrary collections of spatial points, but rather on one-piece regions \cite{cohn_chapter_2008}, i.e., on sets of internally connected points:
\begin{align}
    \textit{SpatialObj}(o) \Rightarrow \textit{ProperSR}(\textit{sReg}(o)) 
\end{align}
To provide a formal definition of the concept of \textit{proper spatial region}, we need first to establish a \textit{spatial frame of reference}. 

\subsubsection{Spatial Frame of Reference}
A spatial object is characterised not only with respect to a spatial region but also in terms of a reference coordinate system, also known as \textit{frame of reference}. A frame of reference consists of an origin point and of a set of directed axes intersecting at the origin, $O$. In particular, modelling the 3D space requires three reference axes $X,Y,Z$. 
Although spatial points and spatial regions exist independently of the frame of reference, the interpretation of these spatial primitives only makes sense in the context of a frame of reference. Once we have defined a reference frame, we can interpret spatial points as \textit{geometrical points}, i.e., as coordinate triples in $\mathbb{R}^3$. Let $GP$ be the set of all geometrical points in the considered space:
\begin{align}
    GP = \{p | p = (x,y,z) \in \mathbb{R}^3\}
\end{align}
The identified frame of reference also has an associated \textit{granularity}, i.e., an infinitesimally small constant $D>0$ in $\mathbb{R}$, which defines the minimum distance for two geometrical points to be considered as distinct entities. Two geometrical points are then said to be \textit{adjacent} iff their geometrical distance is equal to $D$. To compute the distance between two geometrical points, they have to be in the same frame of reference. Let $d(gp,gp')$ be a function which returns a real number indicating the geometric distance between points $gp$ and $gp'$. Then:
\begin{align}
    \textit{Adj}(gp,gp') \Leftrightarrow d(gp,gp') = D
\end{align}
The definition of proper spatial region then follows from the defined notion of adjacency:
\begin{align}
  \textit{ProperSR}(sr) \Leftrightarrow \forall gp [gp\in sr \Rightarrow \textit{Conn}(gp,sr)]  
\end{align}
\begin{multline}
    \textit{Conn}(gp,sr) \Leftrightarrow \forall gp' [gp' \in sr \wedge \\ 
gp' \neq gp] \Rightarrow \textit{ConnP}(gp,gp')
\end{multline}
\begin{multline}
 \textit{ConnP}(gp_1,gp_2)  \Leftrightarrow \textit{Adj}(gp_1,gp_2) \vee \\ 
  \exists gp_3[\textit{Adj}(gp_1,gp_3) \wedge  \textit{ConnP}(gp_3,gp_2)]
\end{multline}
In our model, we assume that the \textit{global spatial region}, $GP$, is a fully-connected set of points.
Moreover, we assume that spatial regions can be approximated through 3D boxes\footnote{\noindent In the case of large regions of negligible thickness, such as floors, walls and ceilings, the spatial region reduces to a 2D surface.}. This simplifying assumption is consistent with standard practice in the literature \cite{deeken_grounding_2018,borrmann_query_2010}. Bounding boxes can have an arbitrary orientation around the Z axis aligned with gravity, but their base is always parallel to the XY plane. In particular, we consider the minimum bounding box which best approximates the real volume occupied by an object and which is aligned with its \textit{natural orientation} \cite{chiatti_towards_2020}. Let $b$ be a set of geometrical points which contains the spatial region of $o$:
\begin{align} 
& \textit{BoundBox}(b,o) \Leftrightarrow    \textit{sReg}(o) \subseteq b \wedge  
   b \subseteq GP 
\end{align}
\begin{multline}
    \textit{MinBoundBox}(b,o)  \Leftrightarrow \textit{BoundBox}(b,o) \wedge 
   \\ 
    \neg \exists b'[\textit{BoundBox}(b',o) \wedge b' \subset b]
\end{multline}
In this scenario, the environment navigated by a robot can also be modelled as a spatial region including an arbitrary number of objects, i.e., as a global spatial region. Consequently, the outer region of a spatial region, $sr$, is:
\begin{align}
&   \textit{outReg}(sr) = \{gp|gp \in GP \wedge gp \notin sr \}
\end{align}
The frame of reference of the global region, $F_g$, is \textit{extrinsic}, i.e., based on a reference point which is external to both an object and an observer. $F_g$ remains fixed as the robot navigates the environment. Conversely, the robot's frame of reference, $F_r$, changes as the robot moves. Thus, it is \textit{deitic}, relative to the observer's position. Consequently, the location of objects at each point in time can be interpreted differently, based on which frame of reference is considered. Within the formal spatial reasoning frameworks of \cite{borrmann_query_2010,deeken_grounding_2018}, all the spatial relations between objects are defined according to the same pre-defined frame of reference, whether it is an extrinsic, deitic or intrinsic one, i.e., inherent to a specific object. 

\begin{figure*}
  \includegraphics[width=\textwidth,trim=0 190 0 25, clip]{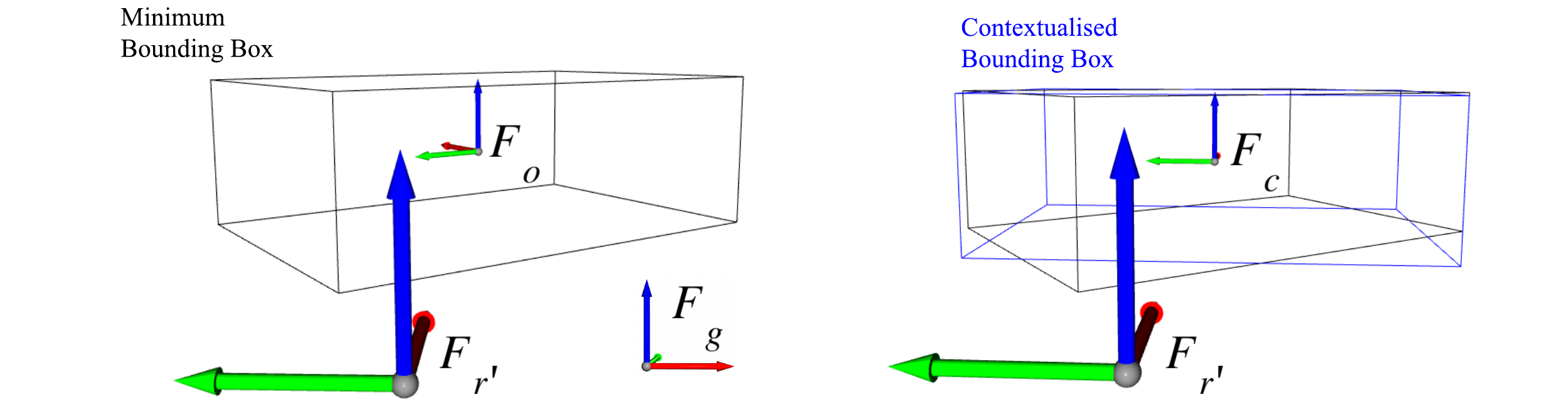}
  \caption{The robot's viewpoint, $F_{r'}$, consists of an origin and three axes $X_{r'}$ (in red), $Y_{r'}$ (in green) and $Z_{r'}$ (in blue). $F_{r'}$ may not coincide with the frame of reference characterising the global map, $F_g$, nor with the intrinsic frame of reference of a certain object, $F_o$. As shown on the left-hand side of the Figure, a spatial object is first modelled as the minimum 3D box bounding the object. Then, $F_{r'}$ is translated to the object's centroid to define a contextualised frame of reference $F_c$. Moreover, a Contextualised Bounding Box (highlighted in blue) is generated, i.e., the bounding box which requires the minimum rotation along the $Z$ axis to align the minimum bounding box with $F_c$. }
  \label{fig:boxes} 
\end{figure*}

Differently from the latter relations, linguistic spatial predicates implicitly refer both to (i) the location of whichever object is considered as reference, and to (ii) the observer's point of view \cite{landau_what_1993}. Similarly, a robot would conclude that ``A is on the left of B'' based not only on the location of objects A and B within $F_g$, but also on $F_r$. From a different standpoint, A might appear on the right of B, for instance, or in front of it.
To model such cases, we introduce the notion of \textit{robot's viewpoint}, $F_{r'}$. Let  $C_{o}$ be the centroid of the spatial region representing object $o$. Then, $F_{r'}$ is obtained by rotating $F_r$ along $Z_r$, by an angle $\alpha$. Specifically, $\alpha$ is the angle between $X_r$ and the imaginary line connecting the origin of $F_r$ with $C_o$.    
Let $F_o$ of origin $C_{o}$ and axes $X_o,Y_o,Z_o$ be the intrinsic frame of reference of $o$, i.e., the frame of reference which is aligned with the orientation of $o$. 
Then, the contextualised frame of reference of the object, $F_c$, is the frame of reference of origin $C_{o}$ whose axes have the same orientation of the axes defining the robot's viewpoint, $F_{r'}$ (Figure \ref{fig:boxes}). 

Based on $F_c$, we can construct a \textit{Contextualised Bounding Box (CBB)}, which is obtained by aligning the minimum bounding box with $F_c$. Let $\textit{rotZ}(b,\theta)$ be a function which returns the spatial region, $sr$, obtained by rotating an input bounding box along $Z$ by an angle $\theta$. Given a frame of reference $F_c$, then $\textit{yaw}(sr,F_c)$ returns the angle between the intrinsic frame of reference of $sr$ and $F_c$, along $Z$. Then, given $\pi/2$:
\begin{multline}
    \textit{IsCBB}(\textit{rotZ}(b,\theta),o) \Leftrightarrow 
    \textit{MinBoundBox}(b,o)  \wedge  \\
    \exists \theta[\textit{mod}(\textit{yaw}(\textit{rotZ}(b,\theta),F_c), \pi/2) = 0 \wedge \neg \exists \theta'  \\
    [\textit{mod}(\textit{yaw}(\textit{rotZ}(b,\theta'),F_c), \pi/2 ) = 0 \wedge \theta' < \theta]] 
\end{multline}
Namely, to construct CBB, we select the minimum angle $\theta$ so that the value returned by the \textit{yaw} function is divisible by $\pi/2$, i.e., the remainder of their division, \textit{mod}, is zero. There are always four possible alignments of a bounding box, $b$, for which \textit{mod} is zero. Thus, by selecting the minimum angle among these four, we apply the transformation which is least disruptive of the natural orientation of the object.
Thanks to these newly-defined spatial concepts, we can now map the metric, topological and directional relations in \cite{borrmann_query_2010,deeken_grounding_2018} to commonsense predicates expressed in natural language.  

\subsubsection{Metric spatial relations} Given two spatial objects $o_1,o_2$ and two geometrical points $gp_1,gp_2$ where $gp_1 \in o_1$ and $gp_2 \in o_2$, we define the distance between two geometrical points as the their Euclidean distance:
\begin{equation}
\resizebox{.9\linewidth}{!}{$
    \displaystyle
    d(gp_1,gp_2) =  \sqrt{(x_1 -x_2)^2 + (y_1 -y_2)^2 + (z_1 -z_2)^2}
$}
\end{equation}
 Then, the distance between two spatial objects is defined as the global minimum of the pointwise distance function, $d$:
\begin{multline}
[\textit{distance}(o_1,o_2) =  d(gp_1,gp_2)] \Leftrightarrow  gp_1 \in o_1 \wedge gp_2 \in o_2 \wedge \\ 
\forall gp_3,gp_4 [gp_3 \in o_1 \wedge 
 gp_4 \in o_2] \Rightarrow d(gp_3,gp_4) > \\d(gp_1,gp_2) 
\end{multline}%
A distance threshold, $T$, can be then introduced, to represent closeness between objects. That is, for a $T$ greater than or equal to the frame granularity $D$ defined earlier:
\begin{align}
    \textit{IsClose}(o_1,o_2) \Leftrightarrow   \textit{distance} (o_1,o_2)  \leq T 
\end{align}
In particular, if the minimum distance between two objects equals $D$, then the two objects touch:
\begin{align}
   & \textit{Touches}(o_1,o_2) \Leftrightarrow  \textit{distance}(o_1,o_2) = D
\end{align}

\subsubsection{Topological spatial relations}
Topological relations are spatial relations which are invariant under a topological isomorphism, i.e., a function $f: X \to Y$ which preserves neighbourhood relationships while mapping $X$ to $Y$. Although several different qualitative representations of topological relations have been proposed \cite{cohn_chapter_2008}, here we focus on a subset of topological relations, namely on the intersection and containment relations. As shown in the remainder of this Section, this minimal subset of relations, combined with metric and directional relations, is sufficient to cover all the commonsense spatial relations required in the scenario of interest. First, based on our prior definitions, two spatial regions, $sr,sr'$ intersect iff they have at least one geometrical point in common:
\begin{align}
    & \textit{Int}(sr,sr') \Leftrightarrow \exists gp [gp \in sr \wedge 
      gp \in sr']
\end{align}
We also define the spatial region representing the intersection between two objects (i.e., the intersection between the associated spatial regions) as follows:
\begin{equation}
\resizebox{.9\linewidth}{!}{$
    \displaystyle
    \textit{inter}(o1, o2) = \{gp | gp \in \textit{sReg}(o1) \wedge gp \in \textit{sReg}(o2)\}
$}
\end{equation}
Then, a special case of the intersection relation is the case where one spatial region completely contains the other: 
\begin{align}
   & \textit{ComplCont}(sr,sr') \Leftrightarrow  \forall gp 
    [gp \in sr'  \Rightarrow 
    gp \in sr] 
\end{align}
Semantically, $o$ contains $o'$ completely iff all the geometrical points in the spatial region of $o'$ are also members of the spatial region of $o$. 

\subsubsection{Directional spatial relations}
Differently from metric and topological relations, directional spatial relations are interpreted differently based on the considered frame of reference. \cite{borrmann_query_2010} have proposed a qualitative representation for directional relations where the region outside a 3D bounding box is partitioned into six halfspaces, i.e., one halfspace for each semi-axis of $X,Y,Z$. Because we have defined a global spatial region containing all geometrical points in the robot's environment, these halfspaces are also proper spatial regions, which can be approximated through 3D bounding boxes. In particular, as in \cite{deeken_grounding_2018}, they can be modelled as 3D extrusions, obtained by multiplying the extent of the object spatial region by a scaling factor $s \in \mathbb{R}$. 

The coordinates of all geometrical points in the minimum bounding box are bound to a minimum and maximum value, e.g., $x_{min}$ and $x_{max}$. Let $X_{o}^{+}$ and $X_{o}^{-}$ be the positive and negative semi-axes of $X_o$ in $F_o$. Then, we define a function, $\textit{hs}$, which returns the halfspace of an input bounding box, given semi-axis, $X_{o}^{+}$, and frame of reference, $F_o$:
\begin{multline}
    \textit{MinBoundBox}(mb_1,o_1) \Rightarrow  \textit{hs}(mb_1,X_{o}^{+}, F_o) = \\
\begin{split}
    \{gp \in \textit{outReg}(mb_1)| &{} gp=(x,y,z) \textit{ w.r.t } F_o,   \\
   & x_{max} \leq x \leq x_{max} + x_{max} \cdot s , \\
   &  y_{min} \leq y \leq y_{max}, \\
   &  z_{min} \leq z \leq z_{max}  \} 
   \end{split}
\end{multline}
Additional halfspaces can be similarly derived for the other semi-axes in $F_o$, as further documented in the supplementary materials. Once these halfspaces have been defined, one can test whether a second object $o_2$ lies within any of the halfspaces of $o_1$. In particular, \cite{borrmann_query_2010} differentiate between ``relaxed'' (\_r) and ``strict'' (\_s) spatial operators, based on whether $o_2$ intersects or is completely contained in the halfspaces of $o_1$. In the following, we represent predicates symbolising cardinal directions \textit{East}, \textit{West}, \textit{North}, \textit{South}, \textit{Above} and \textit{Below} through their capital initial. Given a $F_o$ which coincides with $F_g$, the strict and relaxed 
definitions of the relation \textit{East}($o_2$,$o_1$) are:
\begin{align}
\begin{split}  \textit{E\_s} (o_2, o_1, &{}F_o) \Leftrightarrow \textit{MinBoundBox}(mb_1,o_1) \wedge \\
&  \textit{ComplCont}(\textit{hs}(mb_1,X_{o}^{+}, F_o),\textit{sReg}(o_2)) \end{split}
\end{align}
\begin{multline}
\textit{E\_r} (o_2,o_1,F_o) \Leftrightarrow \textit{MinBoundBox}(mb_1,o_1) \wedge \\ \textit{Int}(\textit{hs}(mb_1,X_{o}^{+}, F_o),\textit{sReg}(o_2))
\end{multline}
Based on our prior definitions, we can model directional relations with respect to a given $F_o$ as follows:
\begin{align}
   & \textit{W\_r} (o_2,o_1,F_o) \Leftrightarrow   \textit{Int}(\textit{sReg}(o_2),\textit{hs}(mb_1,X_{o}^{-},F_o)) \\
   & \textit{N\_r} (o_2,o_1,F_o) \Leftrightarrow   \textit{Int}(\textit{sReg}(o_2),\textit{hs}(mb_1,Y_{o}^{+},F_o)) \\
    &    \textit{S\_r} (o_2,o_1,F_o) \Leftrightarrow   \textit{Int}(\textit{sReg}(o_2),\textit{hs}(mb_1,Y_{o}^{-},F_o)) \\
    & \textit{A\_r} (o_2,o_1, F_o) \Leftrightarrow  \textit{Int}(\textit{sReg}(o_2),\textit{hs}(mb_1,Z_{o}^{+},F_o)) \\
    & \textit{B\_r} (o_2,o_1,F_o) \Leftrightarrow   \textit{Int}(\textit{sReg}(o_2),\textit{hs}(mb_1,Z_{o}^{-},F_o))
\end{align}
For brevity, we have omitted the predicate $\textit{MinBoundBox}(mb_1,o_1)$ from axioms 23-27. The full definition is in the supplementary materials.
\begin{figure*}[t]
  \includegraphics[width=\textwidth,trim=0 25 0 10, clip]{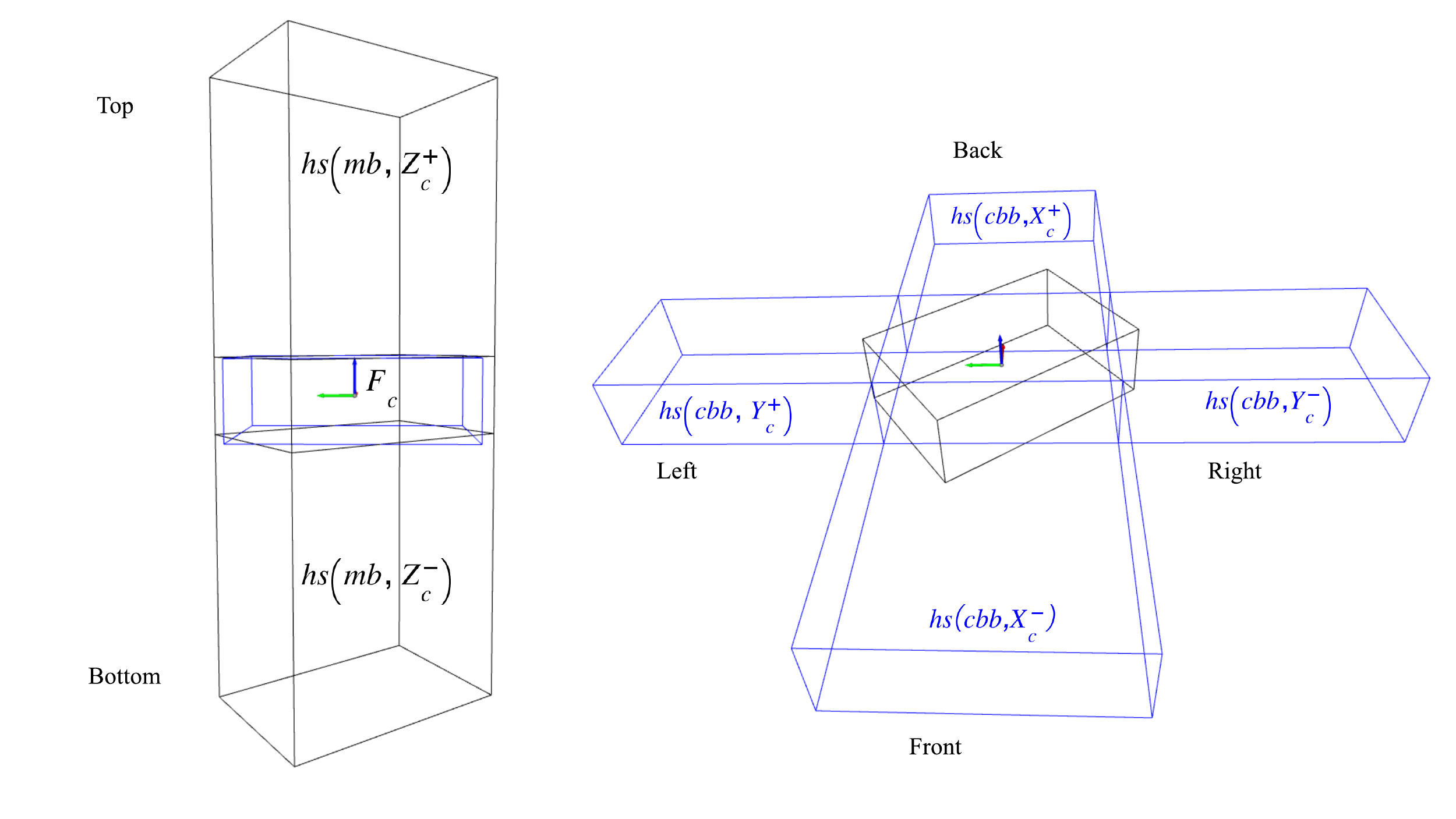}
  \caption{Halfspaces are generated by extruding the 3D bounding boxes along the axis direction in the frame of reference. Specifically, the top and bottom halfspaces, i.e., extruded along the $Z$ axis, are derived from the minimum oriented bounding box (left-hand side of the Figure). The left, right, front and bottom halfspaces are instead extruded from the Contextualised Bounding Box (right-hand side of the Figure).}
  \label{fig:hs} 
\end{figure*}

\cite{borrmann_query_2010} defined the aforementioned relations under the assumption that $F_o$ is always aligned with $F_g$. However, this assumption does not hold in the case of mobile robot sensemaking. Indeed, the frame of reference of the robot, $F_r$ is mobile, i.e., its origin and orientation change over time. Moreover, the natural orientation of objects may not be aligned with $F_g$. 
Thus, to produce a representational model which suits the case of robot sensemaking, we need to map axioms 22-27 to the contextualised frame of reference, $F_c$, defined earlier. 
In a typical robotic setting, the robot always faces towards $X_{r}^{+}$, and $Z_{r}^{+}$ is directed upwards, i.e., opposite to the direction of gravity. Then, the orientation of $Y_{r}$ is given by applying the right hand rule (Figure \ref{fig:boxes}). Based on these premises, all $Z$ axes always share the same orientation. Namely, the top and bottom halfspaces of an object do not change with the robot's reference frame (Figures \ref{fig:boxes},\ref{fig:hs}). Therefore, the \textit{A\_r} and \textit{B\_r} predicates based on the minimum oriented bounding box of an object w.r.t. a given $F_o$ can be directly reused to define the \textit{Above} and \textit{Below} relations w.r.t. $F_c$: 
\begin{align}
& \textit{Above} (o_2,o_1,F_c) \Leftrightarrow \textit{A\_r} (o_2,o_1,F_o) \\
& \textit{Below} (o_2,o_1,F_c) \Leftrightarrow \textit{B\_r} (o_2,o_1,F_o)  
\end{align}
Nonetheless, to model relations such as \textit{RightOf} or \textit{LeftOf}, we need to account for the robot's viewpoint. Thus, we apply the halfspace-based model to the Contextualised Bounding Box we have defined earlier (see Figure \ref{fig:hs}). By definition, CBB is aligned with the contextualised frame of reference, $F_c$, so the front halfspace of CBB, for instance, can be defined w.r.t. a given $F_c$ as follows:
\begin{multline}
    \textit{IsCBB}(cbb_1,o_1) \Rightarrow \textit{hs}(cbb_1,X_{c}^{-},F_c) = \\
\begin{split}
    \{gp \in \textit{outReg}(cbb_1) | gp &{} =(x,y,z) \textit{ w.r.t.} F_c,  \\
  &  x'_{min} - x'_{min} \cdot s \leq x \leq x'_{min}, \\
  &  y'_{min} \leq y \leq y'_{max}, \\
  &  z'_{min} \leq z \leq z'_{max}
    \} \end{split}
\end{multline}
Capitalising on these spatial constructs, we can define the remaining directional relations:
\begin{align}
  &  \textit{RightOf} (o_2,o_1,F_c) \Leftrightarrow  \textit{Int}(\textit{sr}_2,
    \textit{hs}(cbb_1,Y_{c}^{-},F_c)) \\
& \textit{LeftOf} (o_2,o_1,F_c) \Leftrightarrow  \textit{Int}(\textit{sr}_2,
 \textit{hs}(cbb_1,Y_{c}^{+},F_c)) \\
& \textit{InFrontOf}(o_2,o_1,F_c)  \Leftrightarrow  \textit{Int}(\textit{sr}_2,
 \textit{hs}(cbb_1,X_{c}^{-},F_c)) \\
 &\textit{Behind}(o_2,o_1,F_c)  \Leftrightarrow  \textit{Int}(\textit{sr}_2,
 \textit{hs}(cbb_1,X_{c}^{+},F_c)) 
\end{align}
For brevity, in axioms 31-34 we have omitted the predicate $\textit{IsCBB}(cbb_1,o_1)$, which is always valid. The full definition is given in the supplementary materials.
Next, we need to specify how the qualitative spatial relations we have identified align with commonsense spatial predicates. 

\subsubsection{Commonsense spatial relations}
In English, objects are represented by nouns while the spatial relationships between objects are mainly represented through prepositions - e.g., on, next to, behind \cite{landau_what_1993}. Spatial relations are also implied by using certain verbs (e.g., person wears shirt). However, almost invariably, these verbs can be reduced to a simplified form, followed by a preposition (e.g., person has shirt on). Hence, the canonical structure of a spatial sentence consists of three elements: (i) a \textit{reference} object and (ii) a \textit{figure} object, both expressed as noun phrases, as well as (iii) a spatial preposition. The reference object and the preposition, together, define the spatial region occupied by the figure object. \\
As pointed out in \cite{landau_what_1993}, a set of reference axes is needed to differentiate the \textit{front}, \textit{back}, \textit{top}, \textit{bottom} and \textit{sides} of an object. Specifically, the object's \textit{top} and \textit{bottom} are defined as ``the regions at the ends of whichever axis is vertical in the object's normal orientation" \cite{landau_what_1993}. Thus, they are conceptually equivalent to the notion of top and bottom halfspaces we defined for the minimum oriented bounding box. Moreover, the object \textit{front} is defined as the region at the end of the object's horizontal axis which also faces the observer.  Conversely, the object \textit{back} region is located opposite to the observer along the same axis. Finally, the region at the end of any other horizontal axis can be called a \textit{side}. Thus, the four halfspaces we have defined based on the CBB cover these concepts. \\
Consequently, the directional relations defined at statements 28-34 are fit to model the directional predicates in \cite{landau_what_1993}. Moreover, the \textit{LeftOf} and \textit{RightOf} relations can be combined so that, given $F_c$: 
\begin{align}
\begin{split}
   \textit{Beside} (o_2,o_1,F_c) \Leftrightarrow &{} \textit{RightOf} (o_2,o_1,F_c) \lor  \\  & \textit{LeftOf} (o_2,o_1,F_c) \end{split}
\end{align}
An interesting case is that of the ``on" preposition. One of the senses of ``on" is semantically related to ``above". However, while ``above" typically implies absence of contact between the two objects, ``on" strongly favours a contact reading \cite{landau_what_1993}. Formally, we make this distinction by defining:
\begin{align}
\begin{split}
     \textit{OnTopOf} (o_2,o_1,F_c)\Leftrightarrow &{} \textit{Above} (o_2,o_1,F_c)  \wedge  \\ & \textit{Touches} (o_2,o_1) 
     \end{split}
\end{align}
Nonetheless, the ``on" preposition can also be used to denote that the figure object is supported by the reference object. For instance, we say that a ``clock is on the wall" although the two objects overlap horizontally. The phrase ``clock on wall" also implies that the wall is adequately stable to support the clock. Indeed, if two objects differ in terms of size and mobility, we typically prefer to consider the larger and more stable object as reference \cite{landau_what_1993}. To disambiguate these additional uses of ``on", we define, for a given $F_c$:
\begin{align}
    \begin{split} \textit{LeansOn} (&{}o_2,o_1,F_c) \Leftrightarrow \textit{Touches} (o_2,o_1)  \wedge \\ 
    & \neg \textit{Above} (o_2,o_1,F_c) \wedge \neg \textit{Below} (o_2,o_1,F_c) \wedge  \\
    \exists o_3 &[\textit{Touches}(o_2,o_3) \wedge  \textit{Below} (o_3,o_2,F_c)]  \end{split}  
\end{align}
\begin{multline}
    \textit{Touches} (o_2,o_1)\wedge \neg \textit{Above}(o_2,o_1,F_c) \wedge \\
    \neg \exists o_3 \textit{Touches}(o_3,o_2) \Rightarrow \textit{AffixedOn} (o_2,o_1,F_c)
\end{multline}
Namely, whenever $o_2$ is supported by a reference object $o_1$ along the horizontal direction, it is typically said to be ``leaning against" $o_1$: e.g., a ladder leaning against a wall. Furthermore, if the reference object $o_1$ provides the only support surface for $o_2$, $o_2$ is typically said to be \textit{AffixedOn} $o_1$: e.g., a ladder which is affixed on the wall, above ground. Nonetheless, there may be cases where an object, $o_2$, is physically affixed to a surface, $o_1$, even though $o_1$ is not the only surface in contact with $o_2$: e.g., a ladder affixed at ground level. Hence, we used a single logic implication in Statement 37. 

Similar considerations apply for the spatial preposition ``in'', which is also polysemous. First, ``in'' is generally used to imply that one object is ``inside'' another, or, based on our prior topological definitions, that one object is completely contained in the other (axiom 19).
However, ``in" is also used in cases where two objects only partially compenetrate each other. For instance, we would say that ``a cat is in the box" even when the cat's tail is peeping from the box. To define this notion of partial containment we need first to define a function, \textit{adjSRCard}, which, given two spatial regions, sr and sr’, returns the cardinality of the set of points in sr’ that are adjacent to points in sr:
\begin{multline}
\textit{adjSRCard}(sr, sr') =  | \{gp' | gp' \in 
    \textit{outReg}(sr) \wedge gp' \in sr' \wedge \\ \exists gp[gp \in sr \wedge \textit{Adj}(gp,gp')] \} | 
\end{multline}
Hence, we can now define partial containment as follows:
\begin{align}
    & \textit{PartIn}(o_1,o_2) \Leftrightarrow   \textit{sr}=\textit{inter}(\textit{sReg}(o_1), \textit{sReg}(o_2)) \wedge \nonumber \\ & \textit{adjSRCard}(sr, \textit{sReg}(o_1)) <  \textit{adjSRCard}(sr, \textit{sReg}(o_2))] 
\end{align}
Namely, $o_1$ is partially contained in $o_2$ iff the number of points in $o_1$ that are adjacent to the intersection region of $o_1$ and $o_2$ is strictly smaller than the number of points in $o_2$ that are adjacent to the same intersection region. 

Lastly, an object is said to be ``near" another object if it is located in a region ``extending up to some critical distance" \cite{landau_what_1993}. This notion corresponds exactly to our definition of predicate \textit{IsClose} (axiom 15). 

\section{Framework Applicability}
In this Section, we assess the applicability of the proposed logic framework in concrete robotic scenarios. First, we illustrate how the raw sensor data can be opportunely processed to populate a spatial database (Section 4.1). Then, in Section 4.2, we evaluate the extent to which state-of-the-art GIS operators can cover the proposed set of Qualitative Spatial Relations. Furthermore, we show that, once a set of basic spatial concepts has been derived through GIS operators, our framework provides a method to combine these basic spatial concepts to model the commonsense spatial predicates of \cite{landau_what_1993}. 

\subsection{Populating the Spatial Database}
\begin{figure*}[t]
  \includegraphics[width=\textwidth,trim=0 100 0 0, clip]{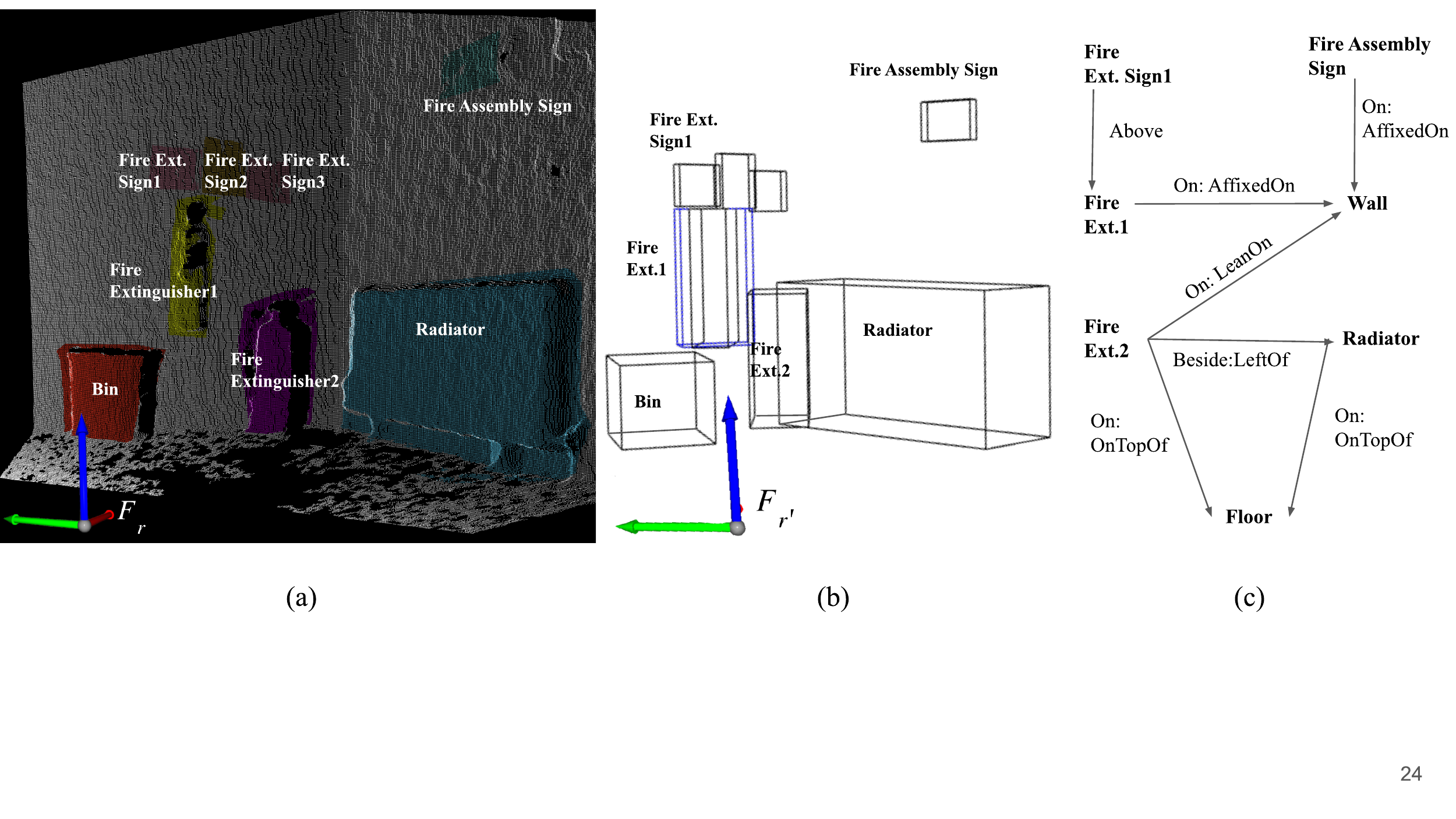}
  \caption{Example of operational workflow: (a) the PointCloud representing the observed scene is first segmented and annotated with object categories. Then, (b) the minimum oriented bounding boxes and CBBs (in blue) are constructed. Lastly, (c) a set of QSR in figure-reference form is derived. In Figures 3b, 3c we show a subset of the bounding boxes and QSR representing the scene, for readability.}
  \label{fig:robot} 
\end{figure*}

Figure \ref{fig:robot}a shows an example of RGB-Depth (RGB-D) data collected through HanS' Orbbec Astra Pro monocular camera. 
At each time frame, $t$, the distance between the robot's pose and the surfaces reached by the laser in the depth sensor is measured. These data are also known as \textit{depth images}, and can be converted to collections of 3D geometrical points in the considered frame of reference, i.e., \textit{PointClouds}. 
As in \cite{chiatti2021aaaimake}, RGB images are autonomously classified through Machine Learning (ML), i.e., based on the multi-branch Network of \cite{zeng2018robotic}. We then project the object regions annotated on RGB images on the PointCloud representing the observed scene, to obtain a segmented 3D region for each annotated object. 
In addition to the annotated object regions, we extract the planar surfaces representing the wall and floor areas. Specifically, we use the PCL library \cite{rusu2011pcl} to reproduce the RANSAC planar segmentation algorithm. Then, we differentiate walls from floors based on the orientation of the plane normal. 
Because the robot pose at each $t$ is known, the depth values within each 3D object region can be converted to coordinate triples in the global frame of reference, $F_g$. In sum, we have produced a semantic map of the robot's environment. 

Consistently with \cite{deeken_grounding_2018}, we store the object regions and labels in the semantic map within a spatial database, implemented in PostgreSQL\footnote{\url{https://www.postgresql.org/}}. By linking these data to a spatial database, we can capitalise on the PostGIS engine\footnote{https://postgis.net/}, which provides a series of query operators for spatial reasoning over PostgreSQL databases. In particular, we rely on the SFCGAL backend\footnote{http://www.sfcgal.org/}, which extends PostGIS by supporting more advanced 3D operations. Objects are stored in the PostGIS database using a minimum oriented polyhedron derived by applying the convex hull algorithm on the segmented PointCloud. Planar surfaces are instead stored as 2D polygons. To populate the spatial database, for each 3D solid or 2D polygon, a new database record is added, which includes: (i) a unique identifier, obtained by concatenating the data collection timestamp with an incremental digit; (ii) the robot's heading and $x,y,z$ coordinates w.r.t. $F_g$; (iii) the top-5 object labels and related confidence scores, as predicted through ML; as well as (iv) a set of Bounding Box representations of the 3D solid, as further detailed in the next Section.  

\subsection{Coverage Study}
The mapping of spatial concepts in our framework to GIS operators is summarised in Table \ref{tab:coverage}. In the following, we also explain the operational steps applied to obtain the output of each row in Table \ref{tab:coverage}.

\subsubsection{Minimum Oriented Bounding Box} To compute the minimum oriented bounding box, we input the 2D projection of the solid on the XY plane to the \textit{ST\_OrientedEnvelope} operator. Second, the \textit{ST\_ZMin} and \textit{ST\_ZMax} functions can be used to find the minimum and maximum coordinate of the 3D solid with respect to the vertical axis. The absolute difference between these two coordinates yields the height of the target bounding box, $h$. Then, the \textit{ST\_Extrude} operator can be used to extrude the 2D envelope along $Z$ by $h$. 
\subsubsection{Contextualised Bounding Box} The Contextualised Bounding Box, CBB, is obtained by rotating the minimum oriented bounding box by an angle $\theta$, to align it with $F_c$ (axiom 12). This operation can be achieved through the \textit{ST\_Rotate} operator. \textit{ST\_Rotate} requires, as input, a geometry, a rotation origin and an angle. We use \textit{ST\_Angle} to compute the angle between the heading of the robot and the bounding box, i.e., $\theta$. To derive the rotation origin, we can exploit \textit{ST\_Centroid}. Since \textit{ST\_Centroid} is a 2D operator, it is applied to the minimum oriented 2D rectangle returned by \textit{ST\_OrientedEnvelope}. Lastly, the transformation is applied to the line passing through the rotation origin and parallel to the $Z$ axis.

\subsubsection{Object Halfspaces} To derive the six object halfspaces, we can rely on the  \textit{ST\_Extrude} operator. 
As emphasised in Table \ref{tab:coverage}, we generate the top and bottom halfspaces by applying \textit{ST\_Extrude} to the minimum oriented bounding box. However, to account for the orientation of both the object and the robot, the remaining four halfspaces are generated by extruding the Contextualised Bounding Box.    
\begin{table*}[t]
    \begin{subtable}[t]{0.45\textwidth}
        \centering
        \begin{tabular}{l | l | l}
        \textbf{Input Geometry/ies} & \textbf{GIS operators applied} & \textbf{Output} \\
        \midrule
        Convex Hull & \textit{ST\_OrientedEnvelope}, & Min Oriented BBox  \\ 
        & \textit{ST\_ZMin}, \textit{ST\_ZMax}, & \\ 
       & \textit{ST\_Extrude} & \\ \hline
       
        Min Oriented BBox, & \textit{ST\_Rotate}, & CBB  \\ 
        Robot heading & \textit{ST\_Angle}, \textit{ST\_Centroid} & \\ \hline 
        Min Oriented BBox, $s$ & \textit{ST\_Extrude} & Top/Bottom Halfspaces\\ \hline
        CBB, $s$ & \textit{ST\_Extrude} & L/R/Front/Back Halfspaces\\ \hline
       Min Oriented BBoxes & \textit{ST\_Volume} & Reference object set\\ \hline
       Min Oriented BBoxes & \textit{ST\_3DDWithin} & \textit{IsClose}, \textit{Touches}\\ \hline
       Min Oriented BBoxes & \textit{ST\_3DIntersects} & \textit{Intersects (Int)}\\ \hline
       Min Oriented BBoxes & \textit{ST\_3DIntersection} & \textit{inter}\\ \hline
       Min Oriented BBoxes & \textit{ST\_3DIntersection}, & \textit{CompletelyContains}\\ 
       &\textit{ST\_Volume} & \textit{(ComplCont)}\\\hline
       Min Oriented BBox & \textit{ST\_3DIntersects} & \textit{Left/RightOf}\\ 
       Halfspaces & & \textit{Above}, \textit{Below} \\
       & & \textit{InFrontOf}, \textit{Behind}\\\hline
       Min Oriented BBoxes & \textit{ST\_Scale}, \textit{ST\_Volume} & \textit{adjSRCard} \\
        & \textit{ST\_Intersection} & \\\hline
       \end{tabular}
       \caption{}
       \label{tab:coverage}
    \end{subtable}
    \hfill
    \begin{subtable}[t]{0.35\textwidth}
        \centering
        \begin{tabular}{l | l  }
        \textbf{QSR} & \textbf{Follows from} \\
        \midrule
        \textit{Beside} & \textit{RightOf}, \textit{LeftOf} \\ \hline 
        \textit{OnTopOf} &  \textit{Touches}, \textit{Above}\\ \hline
        \textit{LeansOn} & \textit{Touches}, \textit{Above} \\ 
        & \textit{Below} \\\hline
        \textit{AffixedOn} & \textit{Touches}, \textit{Above}  \\ \hline
        \textit{Inside} & \textit{ComplCont} \\ \hline
        \textit{PartIn} & \textit{inter}, \textit{adjSRCard}  \\\hline
        \textit{Near} & \textit{isClose} \\ \hline
        \end{tabular}
        \caption{}
        \label{tab:combo}
     \end{subtable}
     \caption{(a) Coverage of spatial notions through PostGIS operators. (b) The basic spatial relations covered by PostGIS are combined to derive more complex QSR.}
     \label{tab:temps}
\end{table*}
\subsubsection{Identifying the reference objects}  Commonsense spatial relations are expressed with respect to a reference object and are asymmetric \cite{landau_what_1993}. For instance, we would say that \textit{a plant is on the floor} (figure-reference form), but we would never say that \textit{the floor is under a plant} (reference-figure form). Therefore, to extract only the QSR which are in figure-reference form, we need first to identify the set of reference objects in each scene. Reference objects are usually the largest and most stable among the observed objects \cite{landau_what_1993}. Hence, by definition, we consider as reference the objects of large area and negligible thickness, which we have modelled as 2D planes, i.e., walls and floors. Moreover, the \textit{ST\_Volume} operator provides a way to measure the size of the 3D bounding boxes. To compute only QSR which are in figure-reference form, we sort objects by volume in descending order. Then, we only compute the QSR between one object and the nearby objects which are smaller than it, if any is found. As a result, in the example depicted in Figure \ref{fig:robot}, QSR such as \textit{wall behind fire extinguisher1} or \textit{floor under fire extinguisher2} would not be extracted. Thus, this design choice also reduces the computational load of extracting QSR for all pairwise combinations of objects.  

\subsubsection{Metric relations} To identify the set of objects which lie nearby a reference object, we can use the \textit{ST\_3DDWithin} operator. This operator returns true if the minimum 3D Euclidean distance between two objects is within a specified threshold. Thus, it is equivalent to our definition of object closeness (axiom 15). Hence, for all object pairs $o'$ and $o$ for which \textit{ST\_3DDWithin} returns true for a specified $T>0$, the relation \textit{IsClose}($o,o'$) also holds. A special case is that of the \textit{Touches}($o,o'$) relation, where $T=0$.  
\subsubsection{Topological relations} The intersection relation we have defined at axiom 17 is directly covered by the  \textit{ST\_3DIntersects} operator. Similarly, \textit{ST\_3DIntersection}, is equivalent to the aforementioned \textit{inter} function (axiom 18).  
Neither PostGIS nor SFCGAL support 3D containment tests. To circumvent this limitation, we derive containment relations by comparing the volume of objects with the volume of their intersection region, through \textit{ST\_Volume}. Namely, if the volume of the intersection region equals the volume of the smaller object, e.g., $o'$, then \textit{ComplCont}($o,o'$).   

\subsubsection{Directional relations} To derive directional QSR, the \textit{ST\_3DIntersects} operator can be applied to the object halfspaces constructed earlier. For instance, in Figure \ref{fig:robot}c, \textit{fire extinguisher2} is on the left of the \textit{radiator}, because it intersects the left halfspace of the \textit{radiator}. Differently from \cite{deeken_grounding_2018}, however, the left halfspace was here defined on a Contextualised Bounding Box, so that the robot's viewpoint is also accounted for.

\subsubsection{Commonsense spatial relations} As shown in Table \ref{tab:coverage}, PostGIS ensures a full coverage of the basic building blocks of our spatial framework. Then, the commonsense relations defined in Section 3 can be seen as a combination of these building blocks (Table \ref{tab:combo}). For instance, having mapped the \textit{LeftOf} and \textit{RightOf} relation to GIS operators (Table \ref{tab:coverage}) we can also conclude whether $o,o'$ are \textit{Beside} one another (Table \ref{tab:combo}). In the case of \textit{PartIn}, we first apply \textit{ST\_Intersection} to obtain the intersection region of $o,o'$, i.e., \textit{inter}$(o,o')$. Then, to approximate the frontier of points which are adjacent to the intersection region, we scale \textit{inter}$(o,o')$ by a $D$, via \textit{ST\_Scale}. In other words, we obtain a region which is infinitesimally larger than the intersection region. Thus, we can use this  region to test, again through \textit{ST\_Intersection}, which sets of points overlap $o$ and $o'$. Specifically, since we are dealing with geometric regions, the cardinality of each point set (axiom 39) is given by the region volume, i.e., via \textit{ST\_Volume}. In sum, the aforementioned operators, also listed in Table 1, cover the logic functions needed to evaluate \textit{PartIn}$(o,o')$. 

Crucially, the introduction of commonsense QSR allows us to disambiguate polysemous spatial prepositions, such as ``in" or ``on". For instance, in Figure \ref{fig:robot}, both fire extinguishers are generically ``on the wall". However, \textit{fire extinguisher 1}, is affixed on the wall, whereas  \textit{fire extinguisher 2}, is also supported by the floor and thus leans on the wall, i.e., it may or may not be affixed on the wall. Thus, compared to the ``on" preposition, the introduced QSR more precisely express the intuitive spatial and physics relations at play.  

\section{Conclusion and Future Work}
In this paper, we have identified a framework for commonsense spatial reasoning which satisfies the concrete requirements of robot sensemaking in real-world scenarios. Differently from prior approaches to qualitative spatial reasoning in robotics, this framework is robust to variations in the robot's viewpoint and object orientation, thus ensuring scalability to many application scenarios. As highlighted by our coverage study, the proposed framework can be fully implemented by capitalising on  state-of-the-art GIS technologies. Moreover, the proposed framework contributes a cognitively-inspired conceptual layer on top of these basic spatial operators, to model commonsense spatial predicates. The resulting linguistic predicates facilitate Human-Robot Interaction, as well as the integration of background spatial knowledge from external resources. As such, the proposed framework contributes to the broader objective of developing Visually Intelligent Agents (VIA), which can reliably assist us with our daily tasks. 

Our future efforts will be focused on evaluating the performance impacts of augmenting ML-based object recognition methods with the proposed commonsense spatial reasoner. In this context, we will also assess the effects of combining commonsense spatial reasoning with the other types of reasoners contributing to the Visual Intelligence of a robot \cite{chiatti_towards_2020}: e.g., size-based \cite{chiatti2021aaaimake}, motion-based and others.   

\appendix
\section{Supplementary Materials}
\subsection{Halfspace Definitions}
The \textit{hs} function we introduced at axiom 20 of the paper can be similarly applied to the other semi-axes in a given $F_o$. Specifically, given an input bounding box, and frame of reference, $F_o$:
\begin{multline}
    \textit{MinBoundBox}(mb_1,o_1) \Rightarrow  \textit{hs}(mb_1,X_{o}^{-}, F_o) = \\
\begin{split}
    \{gp \in \textit{outReg}(mb_1)| &{} gp=(x,y,z) \textit{ w.r.t } F_o,   \\
   & x_{min} - x_{min}\cdot s \leq x \leq x_{min} , \\
   &  y_{min} \leq y \leq y_{max}, \\
   &  z_{min} \leq z \leq z_{max}  \} 
   \end{split}
\end{multline}

\begin{multline}
    \textit{MinBoundBox}(mb_1,o_1) \Rightarrow  \textit{hs}(mb_1,Y_{o}^{+}, F_o) = \\
\begin{split}
    \{gp \in \textit{outReg}(mb_1)| &{} gp=(x,y,z) \textit{ w.r.t } F_o,   \\
   & x_{min} \leq x \leq x_{max} , \\
   &  y_{max}  \leq y \leq y_{max} + y_{max} \cdot s, \\
   &  z_{min} \leq z \leq z_{max}  \} 
   \end{split}
\end{multline}

\begin{multline}
    \textit{MinBoundBox}(mb_1,o_1) \Rightarrow  \textit{hs}(mb_1,Y_{o}^{-}, F_o) = \\
\begin{split}
    \{gp \in \textit{outReg}(mb_1)| &{} gp=(x,y,z) \textit{ w.r.t } F_o,   \\
   & x_{min} \leq x \leq x_{max} , \\
   &  y_{min} -  y_{min} \cdot s \leq y \leq y_{min}, \\
   &  z_{min} \leq z \leq z_{max}  \} 
   \end{split}
\end{multline}

\begin{multline}
    \textit{MinBoundBox}(mb_1,o_1) \Rightarrow  \textit{hs}(mb_1,Z_{o}^{+}, F_o) = \\
\begin{split}
    \{gp \in \textit{outReg}(mb_1)| &{} gp=(x,y,z) \textit{ w.r.t } F_o,   \\
   & x_{min} \leq x \leq x_{max} , \\
   &  y_{min}  \leq y \leq y_{max}, \\
   &  z_{max} \leq z \leq z_{max} + z_{max} \cdot s  \} 
   \end{split}
\end{multline}

\begin{multline}
    \textit{MinBoundBox}(mb_1,o_1) \Rightarrow  \textit{hs}(mb_1,Z_{o}^{-}, F_o) = \\
\begin{split}
    \{gp \in \textit{outReg}(mb_1)| &{} gp=(x,y,z) \textit{ w.r.t } F_o,   \\
   & x_{min} \leq x \leq x_{max} , \\
   &  y_{min}  \leq y \leq y_{max}, \\
   &  z_{min} -  z_{min} \cdot s \leq z \leq z_{min} \} 
   \end{split}
\end{multline}

Similarly, to complete axiom 30 in the main paper, the back, left and right halfspaces are defined as follows, given $F_c$:

\begin{multline}
    \textit{IsCBB}(cbb_1,o_1) \Rightarrow \textit{hs}(cbb_1,X_{c}^{+},F_c) = \\
\begin{split}
    \{gp \in \textit{outReg}(cbb_1) | gp &{} =(x,y,z) \textit{ w.r.t.} F_c,  \\
  &  x'_{max}  \leq x \leq x'_{max} + x'_{max} \cdot s, \\
  &  y'_{min} \leq y \leq y'_{max}, \\
  &  z'_{min} \leq z \leq z'_{max}
    \} \end{split}
\end{multline}

\begin{multline}
    \textit{IsCBB}(cbb_1,o_1) \Rightarrow \textit{hs}(cbb_1,Y_{c}^{+},F_c) = \\
\begin{split}
    \{gp \in \textit{outReg}(cbb_1) | gp &{} =(x,y,z) \textit{ w.r.t.} F_c,  \\
  &  x'_{min}  \leq x \leq x'_{max}, \\
  &  y'_{max} \leq y \leq y'_{max} + y'_{max} \cdot s, \\
  &  z'_{min} \leq z \leq z'_{max}
    \} \end{split}
\end{multline}

\begin{multline}
    \textit{IsCBB}(cbb_1,o_1) \Rightarrow \textit{hs}(cbb_1,Y_{c}^{-},F_c) = \\
\begin{split}
    \{gp \in \textit{outReg}(cbb_1) | gp &{} =(x,y,z) \textit{ w.r.t.} F_c,  \\
  &  x'_{min}  \leq x \leq x'_{max}, \\
  &  y'_{min} - y'_{min} \cdot s\leq y \leq y'_{min} , \\
  &  z'_{min} \leq z \leq z'_{max}
    \} \end{split}
\end{multline}

\subsection{Directional Relations }
The extended definitions of axioms 23-27 in the main paper for a given $F_o$ are:
\begin{multline}
    \textit{W\_r} (o_2,o_1,F_o) \Leftrightarrow \textit{MinBoundBox}(mb_1,o_1) \wedge \\  \textit{Int}(\textit{sReg}(o_2),\textit{hs}(mb_1,X_{o}^{-},F_o))
\end{multline}
\begin{multline}
    \textit{N\_r} (o_2,o_1,F_o) \Leftrightarrow \textit{MinBoundBox}(mb_1,o_1) \wedge \\    \textit{Int}(\textit{sReg}(o_2),\textit{hs}(mb_1,Y_{o}^{+},F_o))
\end{multline}
\begin{multline}
    \textit{S\_r} (o_2,o_1,F_o) \Leftrightarrow \textit{MinBoundBox}(mb_1,o_1) \wedge \\   \textit{Int}(\textit{sReg}(o_2),\textit{hs}(mb_1,Y_{o}^{-},F_o))
\end{multline}
\begin{multline}
    \textit{A\_r} (o_2,o_1, F_o) \Leftrightarrow \textit{MinBoundBox}(mb_1,o_1) \wedge \\  \textit{Int}(\textit{sReg}(o_2),\textit{hs}(mb_1,Z_{o}^{+},F_o))
\end{multline}
    
\begin{multline}
    \textit{B\_r} (o_2,o_1,F_o) \Leftrightarrow  \textit{MinBoundBox}(mb_1,o_1) \wedge \\  \textit{Int}(\textit{sReg}(o_2),\textit{hs}(mb_1,Z_{o}^{-},F_o))
\end{multline}

Moreover, the full definitions of axioms 31-34 in the main paper, given $F_c$, are:

\begin{multline}
  \textit{RightOf} (o_2,o_1,F_c) \Leftrightarrow  \textit{IsCBB}(cbb_1,o_1) \wedge  \\ \textit{Int}(\textit{sr}_2,
    \textit{hs}(cbb_1,Y_{c}^{-},F_c))  
\end{multline}

\begin{multline}
 \textit{LeftOf} (o_2,o_1,F_c) \Leftrightarrow \textit{IsCBB}(cbb_1,o_1) \wedge \\ \textit{Int}(\textit{sr}_2,
 \textit{hs}(cbb_1,Y_{c}^{+},F_c))   
\end{multline}

\begin{multline}
\textit{InFrontOf}(o_2,o_1,F_c)  \Leftrightarrow \textit{IsCBB}(cbb_1,o_1) \wedge \\ \textit{Int}(\textit{sr}_2,
 \textit{hs}(cbb_1,X_{c}^{-},F_c))    
\end{multline}

\begin{multline}
 \textit{Behind}(o_2,o_1,F_c)  \Leftrightarrow \textit{IsCBB}(cbb_1,o_1) \wedge \\ \textit{Int}(\textit{sr}_2,
 \textit{hs}(cbb_1,X_{c}^{+},F_c))   
\end{multline}
\bibliographystyle{kr}
\bibliography{kr-bib}

\end{document}